\documentclass{tlp}

\usepackage{amssymb}
\usepackage{amsmath}
\usepackage{graphicx}
\usepackage{url}

\newcommand{\at}{\mathit{At}}
\newcommand{\gl}{\mathit{GL}}
\newcommand{\Stab}{\mathit{Stab}}
\newcommand{\n}{\mathit{not}\ }
\newcommand{\rk}{\mathit{rk}}
\newcommand{\head}{\mathit{head}}
\newcommand{\body}{\mathit{body}}
\newcommand{\posBody}{\mathit{posBody}}
\newcommand{\lar}{\leftarrow}
\newcommand{\rar}{\rightarrow}

\newtheorem{proposition}{Proposition}[section]

\newtheorem{lemma}{Lemma}[section]
\newtheorem{corollary}{Corollary}[section]

\newtheorem{example}{Example}[section]

\begin{document}

\title[Guarded resolution for Answer Set Programming]
{Guarded resolution for Answer Set Programming}

\author[V.W. Marek and J.B. Remmel]{
V.W. Marek \\
Department of Computer Science, University of Kentucky,\\
Lexington, KY 40506 \email{marek@cs.uky.edu}
\and
J.B. Remmel\\
Departments of Mathematics and Computer Science,\\
University of California at San Diego, \\
La Jolla, CA 92903 \email{jremmel@ucsd.edu }
\thanks{Partially 
supported by NSF grant DMS 0654060.}
}
\submitted {24 April 2009}
 \revised {15 November 2009}
  \accepted{1 January 2010}

\title{Guarded resolution for Answer Set Programming}

\maketitle

\begin{abstract}
We investigate a proof system based on a {\em guarded resolution rule} and show
its adequacy for stable semantics of normal logic programs. As a consequence,
we show that Gelfond-Lifschitz operator can be viewed as a proof-theoretic
concept. As an application, we find a propositional theory $E_P$ whose
models are precisely stable models of programs. We also find a class of 
propositional theories ${\cal C}_P$ with the following properties.
Propositional models of theories in ${\cal C}_P$ are precisely stable models of
$P$, and the theories in ${\cal C}_T$ are of the size linear in the size of $P$.
\end{abstract}

\begin{keywords}
Guarded Resolution,
Proof-theory for Answer Set Programming
\end{keywords}

\section{Introduction}
In this note, we introduce a rule of proof, called {\em guarded unit
resolution}.  Guarded unit resolution is a generalization of a special case of
the resolution rule of proof, namely, 
{\em positive unit resolution}. In positive unit resolution,
one of the inputs is an
atom unit clause. Positive unit resolution is complete for Horn clauses,
specifically, given a consistent Horn theory $T$ and an atom $p$, the atom $p$
belongs to the least model of $T$, $\mathit{lm}(T)$, if and only if  there is a
positive unit resolution proof of $p$ from $T$ \cite{dg84}. 

The modification we introduce in this note concerns {\em guarded atoms} and 
{\em guarded  Horn clauses}. Guarded atoms are strings of the form: $p: \{r_1,
\ldots, r_m\}$ where $p,r_1,\ldots, r_m$ are propositional atoms. Guarded
Horn clauses are strings of the form $p \lar q_1,\ldots,q_n: \{r_1,\ldots,
r_m\}$ again with $p, q_1,\ldots, q_n, r_1,\ldots,r_m$ propositional atoms.

These guarded atoms and guarded rules will be used to obtain a 
characterization of stable models of normal logic programs. There are many
characterizations of stable models of logic programs. 
In fact, in  \cite {li08}, Lifschitz
lists twelve different characterizations of stable models of logic programs. 
The  characterization of stable models that  we present in this paper has a
distinctly proof-theoretic flavor and makes easy to prove some basic 
results on Answer Set Programming such as Fages' Theorem \cite{fa94},
Erdem-Lifschitz Theorem \cite{el03}, and Dung's Theorem \cite{dk89}. 

It should be observed that in \cite{dk89} Dung and Kanchansut consider
so-called quasi-interpretations which, in the formalism of our paper, can be
viewed as collections of guarded atoms. The difference between our approach and
that of \cite{dk89} is that we elucidate the proof theoretic content of the
Gelfond-Lifschitz operator and show how this technique allows for uniform proof
of various results in the theory of stable models of programs.

The outline of this paper is as follows. First, we introduce the definition of 
the guarded resolution rule of proof and then derive its connections with  the
Gelfond-Lifschitz operator \cite{gl88}. Once we do this, we will obtain the
desired lifting of the classical result on the completeness of positive unit
resolution for Horn theories \cite{dg84} to the context of the 
stable semantics of logic programs. In Section \ref{apps}, we show how guarded
resolution proofs can be used to prove various standard results in the theory
of stable models of propositional programs. Finally, in Section \ref{ext}, we
show how the theory developed in this paper can be used to obtain an algorithm
for computation of stable models that does not use loop formulas and runs in
polynomial space in the size of the program.  

\section{Guarded resolution and Stable Semantics}\label{sec:guard}

By a {\em logic program clause},  we mean a string of the form
\begin{equation}\label{C}
C = p \lar q_1, \ldots, , q_n, \n r_1, \ldots, \n r_m.
\end{equation}
A program $P$ is a set of logic program clauses.

We will interpret program clause $C$ given in (\ref{C}) 
as a guarded Horn clause:
\[
g(C) = p \lar q_1, \ldots, , q_n: \{ r_1, \ldots, r_m\}.
\]
We define $g(P) = \{ g(C) : C \in P\}$.
Observe that when we interpret a logic program clause as a guarded Horn
clause, the polarity of atoms appearing negatively in the body of the
programming clause changes in its representation as the guarded Horn clause.
That is,  they occurred {\em negatively} in the body of clause and they now
appear {\em positively} in the guard. By convention, we think of a 
propositional atom as a guarded atom with an empty guard. 

We now introduce our guarded resolution rule as follows. It has two arguments:
the first is a guarded Horn clause and the second is a guarded atom
$q:\{r_1,\ldots,r_n\}$. The guarded atom $q$  must occur in the body of the
guarded Horn clause. The result of the application of the rule is a guarded
Horn clause whose body is the body of the original guarded Horn clause minus
the atom $q$. The guard of the resulting guarded Horn clause  is the union of
the guard of the guarded atom and the guard of the original guarded Horn
clause. Formally, our guarded resolution rule has the following form:

\[
\frac{p \lar q_1, \ldots, , q_n: \{ r_1, \ldots, r_m\} \quad q_j :
\{s_1,\ldots, s_h\}}{ p \lar q_1, \ldots, q_{j-1} , q_{j+1}, \ldots, q_n: \{
r_1, \ldots, r_m, s_1, \ldots, s_h\} }.
\]

Next, we discuss the Gelfond-Lifschitz operator associated with a normal
propositional program. Given a set of atoms $M$ and a normal 
logic program $P$, we first define 
the Gelfond-Lifschitz reduct $P^M$ of $P$.  $P^M$ is constructed 
according to the following two step process. First, if 
$$C = 
p \lar q_1, \ldots, , q_n, \n  r_1, \ldots, \n r_m$$  
is a clause in $P$ and $r_j \in M$ for some $1 \leq j \leq m$, then 
we eliminate $C$. Second, if $C$ is not eliminated after 
step 1, then 
we replace $C$ by 
$$p \lar q_1, \ldots ,q_n.$$
Clearly, $P^M$ is a
Horn program. Thus $P^M$ has a least model $N_M$. The Gelfond-Lifschitz
operator assigns to $M$ the set of atoms $N_M$.

Our guarded unit resolution rule naturally leads to the notion of a guarded
resolution proof $\cal P$ of a guarded atom $p: S$ from the program $P$. Here
$S$ is a, possibly empty,  set of atoms. That is, a guarded resolution proof of
$p:S$ is a labeled tree 
such that every node that is not a leaf has two parents, one labeled with
a guarded Horn clause and the other labeled with a guarded atom, where 
the label of
the node is the result of executing the guarded unit resolution rule 
on the labels of the
parents. Each leaf is either a guarded Horn clause 
$p \lar q_1, \ldots, , q_n: \{ r_1, \ldots, r_m\}$ such that 
$p \lar q_1, \ldots, , q_n, \n r_1, \ldots, \n r_m$ is in $P$ or 
a guarded atom $q:\{ r_1, \ldots, r_m\}$ such that 
$q \lar \n r_1, \ldots, \n r_m$ is in $P$.  In the special case where 
the tree consist of a single node, we assume that the 
node is labeled with a guarded atom $q:\{ r_1, \ldots, r_m\}$ where 
$q \lar \n r_1, \ldots, \n r_m$ is in $P$. Note that in 
a guarded resolution proof, guards only grow as we proceed down 
the tree. That is, as we resolve, the guards 
are summed up. For this reason, the guard of the root of the proof contains the
guards of {\em every} label in the tree. 

We say that a set of atoms $M$ {\em admits} a guarded atom $p: S$, if $M
\cap S = \emptyset$  and that $M$ admits a guarded resolution 
proof $\cal P$ if it admits
the label of the root of $\cal P$. The following statement follows from the 
containment properties of guards in  a guarded resolution proof.

\begin{lemma}\label{l.1.1}
If $M$ admits the guarded resolution 
proof $\cal P$, then $M$ admits every guarded atom
occurring as a label in $\cal P$ and $M$ is disjoint from the guard of
every guarded clause in $\cal P$.  
\end{lemma}

We then have the following proposition.
\begin{proposition}\label{p.1.1}
Let $P$ be a propositional logic program and let $M$ be a set of atoms.
Then $\mathit{GL}_P(M)$ consists exactly of atoms $p$ such that there 
exists a set of atoms $Z$ where 
the guarded atom $p:Z$ is the conclusion of a guarded
resolution proof  $\cal P$ from $g(P)$ admitted by $M$.
\end{proposition}
Proof: Let $Q = P_M$ and assume that $p \in \mathit{GL}_P(M)$. Then by 
definition, $p \in T^\omega_Q$ where $T_Q$ is the standard one-step 
provability operator for $Q$.  
We claim that we can prove by induction on $n \in N$ that whenever $p \in
T^n_Q$, then there exists a set of atoms $Z$ such that 
$p:Z$ possesses a guarded resolution proof from $g(P)$ 
admitted by $M$. If $n = 1$, then it must be the case that the $p \lar$ 
belongs to $Q$. But then, for some $r_1, \ldots, r_m$,
\[
p \lar  \n r_1, \ldots, \n r_m
\]
belongs to $P$ and $\{r_1,\ldots, r_m\} \cap M = \emptyset$. Therefore the guarded atom $p:
\{r_1,\ldots, r_m\}$ is admitted by $M$ and it possesses a guarded resolution
proof from $g(P)$, namely, the one that consists of the root labeled by $p:\{r_1, \ldots, r_m\}$.  Now, let us assume $p \in
T^{n+1}_Q$. Then there is a clause $C = p \lar q_1,\ldots, q_s$ in $Q$ 
such that $q_i \in T^n_Q$ for $i=1, \ldots, s$. Thus by induction, there 
are sets of atoms 
$S_i$, $1 \le i \le n$, such that $q_i : S_i$ possesses a guarded resolution 
proof from $g(P)$ admitted by $M$.
As $p \lar q_1, \ldots, q_n$ belongs to $Q$, there must exist atoms $r_1,\ldots, r_m \notin M$ such that 
\[
p \lar q_1, \ldots, q_n, \n r_1, \ldots, \n r_m  
\]
is a clause in $P$. It is easy to combine the guarded resolution 
proofs  of $q_i:S_i$,
$1 \le i \le n$ and the guarded 
clause $p \lar q_1,\ldots, q_n : \{r_1,\ldots, r_m\}$
to obtain a guarded resolution proof from $g(P)$ of the following guarded atom: 
\[
p :  S_1 \cup \ldots \cup S_n \cup \{r_1,\ldots, r_m\}.
\]
As all the sets occurring in the guard of this guarded atom are disjoint from
$M$, the resulting guarded resolution proof  is admitted by $M$. This shows the
inclusion $\subseteq$.

Conversely, suppose $p :S$ has a guarded resolution proof $\cal P$ 
from $g(P)$ admitted by $M$.  By the lemma, all the guards occurring in $\cal
P$ are disjoint from $M$. We can then prove by induction on the height of the 
tree $\cal P$ that $p \in \mathit{GL}_P(M)$.  If the height 
of $\cal P$ is $0$, then it must be the case that 
\[
p \lar  \n r_1, \ldots, \n r_m
\]
belongs to $P$ where $S = \{r_1, \ldots, r_m\}$. But since $M \cap S =
\emptyset$, the clause $p \lar$ belongs to $Q$. Hence $p \in \mathit{GL}_P(M)$.

Now, for the inductive step, assume that whenever $q : S$ has a guarded
resolution proof from $g(P)$ that is admitted by $M$ of height less than or equal to $n$, then $q \in
\mathit{GL}_P(M)$. We now show 
that the same property holds for all guarded atoms $p : U$ which 
have a guarded resolution proof from $G(P)$ that is admitted by $M$ 
of the height $n +1$. 
What does 
such a guarded resolution proof look like? First the root must be the result 
of a guarded unit resolution of the form 
\[
\frac{p \lar q : Z_1 \quad q : S_0}{p : Z_1 \cup S_0}.
\]
As $(Z_1 \cup S_0) \cap M = \emptyset$, $Z_1 \cap M = \emptyset$ and $S_0 \cap
M = \emptyset$. Now, $q :S_0$ has a guarded resolution proof from 
$g(P)$ that is admitted by $M$ 
 of height at most $n$ and, hence, $q \in \mathit{GL}_P(M)$ by our 
inductive assumption. Since as we progress down the proof tree, 
 the body of guarded clauses only get smaller  and the guards 
of guarded clauses only get bigger, there 
must exist a path starting at the guarded clause $p \lar q : Z_1$ 
which consists of guarded clauses of the form 
\begin{eqnarray*}
&& p \lar q,q_1, \ldots, q_t:Z_{t+1}\\
&&\vdots \\
&&p \lar q,q_1: Z_2 \\
&&p \lar q : Z_1
\end{eqnarray*}
where $Z_{t+1} \subseteq Z_t \subseteq \cdots \subseteq Z_1$ and 
for each $i$, there is a node in the tree of the form 
$q_i:S_i$ such that the resolution of $p \lar q,q_1, \ldots, q_i:Z_{i+1}$ 
and $q_i:S_i$ results in the clause $p \lar q,q_1, \ldots, q_{i-1}:Z_{i}$.
Now each $q_i:S_i$ is the root of a guarded resolution proof 
from $g(P)$ that is admitted by $M$ 
of height less than or equal to $n$ and, hence, $q_i$ is in $GL_P(M)$ 
for $i =1, \ldots, t$. 

Since $p \lar q,q_1, \ldots, q_p:Z_{t+1}$ is a leaf, 
there must be a clause 
$$p \lar q,q_1, \ldots, q_t, \n r_1, \ldots, \n r_m$$
in $P$ where $Z_{t+1} = \{r_1, \ldots, r_m\}$. Since $M$ admits the proof tree,
it must be the case that $\{r_1, \ldots, r_m\} \cap M = \emptyset$ and, hence, 
$p \lar q,q_1, \ldots, q_t$ is in $Q$. But then, since $q,q_1, \ldots, q_t$ are
in $GL_P(M)$, it follows that $p \in GL_P(M)$. $\hfill\Box$

Proposition \ref{p.1.1} tells us that the Gelfond-Lifschitz operator
$\mathit{GL}$ is, essentially, a proof-theoretic construct. Here is one
consequence, this time a semantic one.

\begin{corollary}\label{c.1.1}
Let $P$ be a propositional logic program and let $M$ be a set of atoms. Then
$M$ is a stable model of $P$ if and only if 
\begin{enumerate}
\item for every $p \in M$, there is a set of atoms $S$ such that 
there is a guarded resolution proof of $p:S$
from $g(P)$ admitted by $M$  and
\item for every $p \notin M$, there is no set of atoms $S$ such that 
there is a guarded resolution proof of $p:S$ from $g(P)$ admitted by $M$.
\end{enumerate}
\end{corollary}

When $P$ is a Horn program Corollary \ref{c.1.1} reduces to the classical
fact \cite{dg84} that the elements of the least model of
the Horn programs are precisely those that can be proved out of clausal
representation of $P$ using positive unit resolution.

Given a finite set of atoms $S$, we write $\neg S$ for the conjunction
$\bigwedge_{q \in S} \neg q$. Next,
let us call $S$ such that $p:S$ has a guarded resolution proof from 
$g(P)$ a {\em
support} of $p$ with respect to $P$.  We can then form an {\em equation} for
$p$ with respect to $P$, $\mathit{eq}_P(p)$, as follows: 
\[
p \Leftrightarrow (\neg S_1 \lor \neg S_2 \lor \ldots  )
\]
where $S_1, S_2,\ldots $ is the list of all supports of $p$ with respect to
$P$. If the only support of $p$ is the empty set, then we let 
$\mathit{eq}_P(p) =p$ and if there are no supports for $p$, then 
we let $\mathit{eq}_P(p) = \neg p$.
Next, let $E_P$ be the propositional theory consisting of
$\mathit{eq}_P(p)$ for all $p \in \mathit{At}$. We then have the following
theorem resembling Clark's completion theorem, except we get it for stable
models, not supported models.

\begin{proposition}\label{p.1.2}
Let $P$ be a propositional program and let $M$ be a set of atoms. Then $M$ is a
stable model of $P$ if and only if $M \models E_P$.  
\end{proposition}
Proof: First, assume that $M$ is a stable model of $P$. Then if 
$p \in M$, it follows from 
Corollary \ref{c.1.1} that there is an $S$ such that 
$p:S$ has a guarded resolution proof admitted by $M$. Hence 
$M \cap S = \emptyset$ and $M \models
\neg S$. Thus $M$ satisfies both $p$ and one of the disjuncts 
on the right-hand side of
$\mathit{eq}_P(p)$. Hence $M \models \mathit{eq}_P(p)$. Next assume that $p 
\notin M$.  Then there is no $Z$ such that $p:Z$ has a guarded 
resolution proof 
 admitted by $M$. It follows that either $\mathit{eq}_P(p)$ equals 
$\neg p$ or $M$ satisfies both negation of $p$ and of the negation 
of every disjunct on the right-hand-side of
$\mathit{eq}_P(p)$. Thus again $M \models \mathit{eq}_P(p)$.  This shows
``if'' part of the theorem. 

For the other direction, suppose that  $M \models \mathit{eq}_P(p)$. 
Then if $p \in M$, either $\mathit{eq}_P(p) =p$ or 
$\mathit{eq}_P(p) = p \Leftrightarrow (\neg S_1 \lor \neg S_2 \lor \ldots  )$. 
In the former case, this means that the tree consisting 
of a single node $p:\emptyset$ is a guarded resolution proof and, hence, 
$p \lar$ is a clause in $P$. Thus $p$ must be in $M$.  
 In the latter case, 
there must be some $S_i$ such that $M \models \neg S_i$.  But 
by definition, $p:S_i$ is the root of some guarded resolution 
proof $\cal P$ for $g(P)$ and 
since every guard in such a guarded resolution proof 
 is contained in $S_i$, it must be the case that  
$M$ admits $\cal P$. But then we have shown that 
$p \in GL_P(M)$.  Thus $M \subseteq GL_P(M)$.  

On the other hand, 
if $p \notin M$, then either $\mathit{eq}_P(p) =\neg p$ or 
$\mathit{eq}_P(p) = p \Leftrightarrow (\neg S_1 \lor \neg S_2 \lor \ldots  )$.
In the former case, there must be be no guarded resolution proofs 
of $p$ and, hence, $p$ is not in $M$. In the latter case, it 
must be that $M$ does not satisfy any $\neg S_i$. This means that 
there is no guarded resolution proof from $g(P)$ whose root is of the 
form $p:S$ such that $M$ admits 
$p$ and, hence, $p \notin GL_M(P)$. This implies 
$GL_P(M) \subseteq M$ and, hence, $GL_p(M) =M$. Thus $M$ is a stable 
model of $P$. $\mbox{\ }\hfill\Box$


If we look carefully at the structure of any resolution proof tree 
of a guarded atom $p:S$, we see that $S$ collects a set atoms 
which guarantee that $p \in N_M$ whenever $M \cap S = \emptyset$. 
Thus in defining $\mathit{eq}_P(p)$, we essentially unfold the atoms in $S$ to
conjunctions of negated atoms $\neg S$ (cf. \cite{bd99}).

We observe that, in principle, when the program $P$ is infinite, the theory
$E_P$ may be infinitary. Specifically, the formulas $\mathit{eq}_P(p)$ may be
infinitary formulas, since the disjunction on the right-hand-side of the
equivalence may be over an infinite set of finite conjunctions. But the
semantics for infinite propositional logic is well-known \cite{ka64} and can be
applied here. The authors studied the necessary and sufficient conditions for
$E_P$ to be finitary in \cite{mr10}.

\section{Some applications} \label{apps}

In this section we will use the results of Section \ref{sec:guard} to derive
the result of Erdem and Lifschitz \cite{el03}. This result generalizes a
theorem by Fages \cite{fa94} which is useful as a preprocessing step 
for the 
computation of stable models of programs. We will also prove a result of
Dung \cite{dk89} on stable models of purely negative programs.

We say that a set of atoms $M$ {\em has levels} with respect to a program $P$ 
if

\begin{enumerate}
\item $M$ is a model  of $P$, and
\item There is a function $\rk : M \rar \mathit{Ord}$ such that, for every $p 
\in M$, there is a clause $C$ such that
\begin{enumerate}
\item $p = \mathit{head}(C)$,
\item $M \models \mathit{body}(C)$, and 
\item For all $q \in \mathit{posBody}(C)$, $\rk (q) < \rk (p)$.
\end{enumerate}
\end{enumerate}

Clearly, the least model of a Horn program has levels since the function
which assigns to an atom $p\in M$, the least integer $n$ such that $p \in
T_P^n(\emptyset)$ is the desired rank function for condition (2).

We now prove the following proposition.
\begin{proposition}\label{p.2.1}
Let $P$ be a propositional logic program and $M \subseteq \at$. Then $M$ is a
stable model of $P$ if and only if  $M$ has levels with respect to $P$.
\end{proposition}
Proof: Clearly, when $M$ is a stable model of $P$, then $M$ has levels with
respect to $P$. Namely, the rank function whose existence is postulated in (2)
is the rank function inherited from the Horn program $P_M$.

Converse implication can be proved in a variety of ways. Our proof, in the
spirit of the proof-theoretic approach of this paper, uses guarded
resolution. That is, assume that $M$ has levels with respect to $P$ 
where $\rk$ is the rank function in condition (2). 
Our goal is to prove that $M = \mathit{GL}_P(M)$. First, let us observe that
since $M \models P$, $\mathit{GL}_P(M) \subseteq M$. Thus, all we need to
show is that $M \subseteq
\mathit{GL}_P(M)$. By Corollary \ref{c.1.1}, we need only show 
that for given any
$p\in M$, there is a $Z$ such that $p:Z$ possesses a guarded resolution 
proof from $g(P)$ that is admitted by $M$.
We construct the desired set $Z$ and guarded resolution proof  
by using the rank function $\rk$. Let $S = \{ \rk (p) : p \in M\}$,
i.e. $S$ is the range of rank function. 
We proceed by transfinite induction on
the elements of $S$.  Let $p$ be an atom in $M$ such that $\rk (p)$ is the
least element of $S$. By assumption, 
there must exist a clause $C$ in $P$  such that $M \models
\body (C)$, $p = \head(C)$ and for all $q\in \posBody (C)$, $\mathit{rk}(q) <
\rk(p)$. Since $M \models \body (C)$, there can be no 
$q$'s in positive part of the body of $C$ because any such $q$ must be in 
$M$ and have rank strictly less than $p$. Thus the 
 clause $C$ has the following form:
\[
p \lar \n r_1,\ldots \n r_m.
\]
As $M \models \body (C)$, $r_1,\ldots, r_m \notin M$. But then
$p:\{r_1,\ldots,r_m\}$ is a guarded atom admitted by $M$ and so 
$p:\{r_1,\ldots,r_m\}$ has a guarded resolution proof from $g(P)$ which 
consists of a single node labeled with $p:\{r_1,\ldots,r_m\}$.

Now, assume that whenever $\beta \in S$, $\beta < \alpha$, and $\rk
(q) = \beta$, then there is a guarded resolution proof of $q:S$ from $g(P)$
admitted by $M$ for some set $S$. Let us assume that $p \in M$ and $\rk (p) =
\alpha$. By our assumption, there is a clause
$C$
\[
p \lar q_1, \ldots, q_n, \n r_1,\ldots \n r_m
\]
with $M \models \body (C)$ and $\rk (q_1) < \rk (p), \ldots, \rk (q_n) < \rk
(p)$. By inductive assumption, for every $q_i$, $1\le i \le n$, there is a
finite set of atoms $Z_i$ such that there is
guarded resolution proof ${\cal D}_i$ from $g(P)$ 
of $q_i: Z_i$ admitted by
$M$. In particular $Z_i \cap M = \emptyset$. We can then 
easily combine the guarded resolution proofs for $q_i: Z_i$ with 
$n$ applications of guarded unit resolution starting with the  
leaf 
\[
p \lar q_1,\ldots, q_n : \{r_1,\ldots, r_m\}
\]
to produce a guarded resolution proof of 
\[
p: Z 
\]
from $g(P)$ where $Z = Z_1\cup\ldots \cup Z_n \cup \{r_1,\ldots,r_n\}$. Since all $Z_i$s
are disjoint from $M$ and $M \cap \{r_1,\ldots, r_m\} = \emptyset$, it 
follows that 
$M \cap Z = \emptyset$. Thus the resulting resolution proof  is admitted by
$M$.  This completes the inductive argument and thus the proof of the
Proposition.  $\hfill\Box$

We observe that, in fact, it is sufficient to limit the functions $\rk$ 
in the definition of $M$ having levels respect to
$P$ to those rank functions 
that take values in $N$, the set of natural numbers.

We get, as promised,  several corollaries. One of these is the result 
of Erdem and Lifschitz
\cite{el03}. Following \cite{el03}, we say that a program $P$ is {\em tight  on
a set of atoms $M$} if there is a rank function $\rk$ defined on $M$ such that
whenever $C$ is a clause in $P$ and $\head (C) \in M$, then for all $q \in
\posBody (C)$, $\rk (q) < \rk (\head (C)).$ Here is the result of Erdem and
Lifschitz. 
\begin{corollary}[Erdem and Lifschitz]\label{c.2.1}
If $P$ is tight on $M$ and $M$ is a supported model of $P$, 
then $M$ is a stable model of $P$.
\end{corollary}
Proof: Indeed, tightness on $M$ requires that for any $p \in M$, 
there is a support for
$p$ and that {\em all} clauses $C$ that provide the support for the presence of
$p$ in $M$  have the property that the atoms in the positive part of the body
of $C$ have smaller rank. In Proposition \ref{p.2.1}, we showed 
that it is enough
to have just one such clause. Since tightness on $M$ implies existence of such
a supported clause, the corollary follows.$\hfill\Box$

Since all stable models are supported \cite{gl88}, one can express
Erdem-Lifschitz Theorem in a more succinct way.
\begin{corollary}[Erdem-Lifschitz]\label{c.2.2}
If for every supported model $M$ of a program 
$P$, $P$ is tight on $M$, then the
classes of supported and stable models of $P$ coincide.
\end{corollary}
Fages Theorem \cite{fa94} is a weaker version of Corollary \ref{c.2.1} (but
with assumptions that are easier to test). Specifically, we say that a program
$P$  is {\em tight} if there is a rank function $\rk$ such that for every
clause $C$ of $P$, the ranks of the atoms occurring in the positive part of the
body of $C$ are smaller than the rank of the head of $C$. Clearly, if $P$ is
tight, then $P$ is tight on any of its models $M$ since $rk$ will also 
witness that $P$ is tight on $M$. Thus one has the following corollary.
\begin{corollary}[Fages]\label{c.2.3}
If $P$ is tight, then the classes of supported and stable models of $P$
coincide.
\end{corollary}
Tightness is a syntactic property that can be checked in polynomial time by 
inspection of the positive call-graph of $P$. This is not the case for the
stronger assumptions of Proposition \ref{p.2.1} and Corollary \ref{c.2.2}.

Let $\Stab (P)$ be the set of all stable models of $P$.  We say that programs
$P$, $P'$ are {\em equivalent} if $\Stab (P) = \Stab(P')$. This notion and its
strengthenings were studied by ASP community \cite{lpv01}, \cite{tr06}.
We have the following fact.
\begin{lemma}\label{l.2.1}
If $P,P'$ prove the same guarded atoms, then $P$ and $P'$ are equivalent.
\end{lemma}

As a corollary we get the following result due to Dung \cite{dk89}
\begin{corollary}[Dung]\label{c.2.4}
For every program $P$, there is purely negative program $P'$ such that $P$,
$P'$are equivalent.  
\end{corollary}

The program $P'$ is quite easy to construct. That is, for each atom $p$, if
$$\mathit{eq}_P(p) = p \Leftrightarrow (\neg S_1 \lor \neg S_2 \lor \ldots
),$$
then we add to $P'$, all clauses of the form 
$$p \lar \n r_{i,1}, \ldots, \n r_{i,{m_i}}$$
where $S_i = \{r_{i,1}, \ldots,  r_{i,{m_i}}\}$. 
If $\mathit{eq}_P(p) = p$, then we add $p \lar$ to $P'$. Finally, 
if $\mathit{eq}_P(p) = \neg p$, then we add nothing to $P'$.  
It is then easy to see 
that $E_P = E_{P'}$ and hence $P$ and $P'$ are equivalent. 
$\hfill\Box$

\section{Computing stable models via satisfiability, but without 
loop formulas or defining equations}\label{ext}
Proposition  \ref{p.1.2} characterized the stable models of
a propositional program in terms of the collection of all propositional
valuations of the underlying set of atoms. In this section, we  give
an alternative characterization in terms of the models of suitably chosen
propositional theories.  The proof of this characterization uses Proposition
\ref{p.1.2}, but relates stable models of finite propositional programs $P$ to
models of theories of size $O(|P|)$.  This is in contrast to Proposition
\ref{p.1.2} since the set of defining equations is, in general, of size
exponential in $|P|$.

A {\em subequation} for an atom $p$ is either a formula $\neg p$ or a formula
\[
p \Leftrightarrow \neg S
\]
where $S$ is a support of a guarded resolution proof
 of $p$ from $P$. Here if $S =
\emptyset$, then by convention we interpret $p \Leftrightarrow \neg S$ to be
simply the atom $p$. The idea is that a subequation either asserts absence of
the atom $p$ in the putative stable model  or provides the reason for the
presence of $p$ in the putative stable model.  A {\em candidate theory} for
program $P$ is the union  of $P$ and a set of subequations, one for each $p \in
\at$. By ${\cal C}_P$ we denote the set of candidate theories for $P$.

\begin{proposition}\label{p.subeq}
\begin{enumerate}
\item Let $T$ be a candidate theory for $P$ (i.e. $T\in {\cal C}_P$). If $T$ is
consistent, then every propositional model of $T$ is a stable model for $P$.
\item For every stable model $M$ of $P$, there is theory $T \in {\cal C}_P$ 
such that $M$ is a model for $T$.
\end{enumerate}
\end{proposition}
Proof: (1) Let $T$ be a candidate theory for $P$ and suppose that 
$M \models T$. We need to
show that $M$ is a stable model for $P$. In other words, we need to show that
\[
\gl_P(M) = M.
\]
The inclusion $\gl_P(M) \subseteq M$ follows from the fact that $M$ is a model
of $P$. That is, since $M$ is a model of $P$, it immediately follows that $M$
is a model of the Gelfond-Lifschitz transform of $P$, $P_M$. Since 
$\gl_P(M)$ is the unique minimal model of $P_M$, it automatically follows 
that $\gl_P(M) \subseteq M$. 

To show that $M \subseteq \gl_P(M)$, suppose that $p \in M$. Then the
subequation for $p$ that is in $T$ can not be $\neg p$. Therefore it is of the
form 
\[
p \Leftrightarrow \neg U_p
\]
where there is some guarded resolution proof $\mathcal{P}$ of $p:U_p$ from 
$g(P)$. Since
$M \models T$ and $p \in M$, $M \models \neg U_p$.  But then 
$M \cap U_p =\emptyset$ so that $M$ admits $\mathcal{P}$. Hence 
by Corollary \ref{c.1.1},  $p \in \mathit{GL}_P(M)$.\\
\ \\
(2)  Let $M$ be a stable model of $P$.  We construct a candidate theory $T$ so
that $M$ is a model of $T$. To this end, for each $p \notin M$, we select $\neg
p$ as a subequation for $p$. For each $p \in M$, we select a guarded 
resolution proof of some $p:U_p$ from $g(P)$ that admitted by $M$.
We then add to $T$ the formula
\[
p \Leftrightarrow \neg U_p.
\]
Clearly, $T$ is a candidate theory, and $M \models T$, as desired. $\hfill\Box$

Next we give an example of our  
approach to reducing the computation of of stable models to
satisfiability of propositional theories. It will be 
clear from this example that our approach avoids having to 
compute the completion of the program and 
thus significantly reduces the size of the input theories.

\begin{example}
{\rm 
Let $P$ be a propositional program as follows:
\begin{eqnarray*}
&&p \leftarrow t, \neg q\\
&&p \leftarrow \neg r\\
&&q \leftarrow \neg s\\
&&t \leftarrow 
\end{eqnarray*}
Let us observe that the atom $p$ has two inclusion-minimal supports, 
namely $\{q\}$  and $\{r\}$. The atom $q$ has just one support, 
namely $\{s\}$, 
and the atom $t$ also has just one support, namely $\emptyset$. 
The atoms $r$ and $s$ have no support at all.

Thus there are {\em three} subequations for $p$: 
\begin{eqnarray*}
&&p \Leftrightarrow \neg q\\
&&p \Leftrightarrow \neg r\\
&&\neg p
\end{eqnarray*}
Now, $q$ has only two subequations: $q \Leftrightarrow \neg s$, and $\neg q$. 
Similarly, $t$ has only  two subequations, $t$ and $\neg t$, but the second one
automatically leads to contradiction whenever it is chosen. Finally each of $r$
and $s$ have just one defining equation, $\neg r$, and $\neg s$, respectively.

First let us choose for $p$, the subequation $\neg p$, and for $q$, the
subequation $q \Leftrightarrow \neg s$. The remaining subequations are forced
to  $t$, $\neg r$, and $\neg s$. The resulting theory has nine clauses, when 
we write our program in propositional form:
$$S = \{ \neg p, \neg r, \neg s, t, q\Leftrightarrow \neg s\} \cup 
\{\neg t \lor p \lor q, r \lor p, s \lor q, t\}.$$
It is quite obvious that this theory is inconsistent. However, 
if we choose for $p$, the subequation $p\Leftrightarrow \neg r$ and 
for $q$, the subequation $q\Leftrightarrow \neg s$, then the resulting 
theory written out in propositional form is 
\[
S = \{p\Leftrightarrow \neg r, \neg r, \neg s, t, q\Leftrightarrow \neg s\}
\cup \{\neg t \lor p \lor q, r \lor p, s \lor q, t\}.
\]
In this case, $\{p,q,t\}$ is a model of $S$ and hence, $\{p,q,t\}$ is a stable
model of $P$. $\mbox{ }\hfill\Box$
}
\end{example}

It should be clear that Proposition \ref{p.subeq} implies an algorithm for
computation of stable models. Namely, we generate candidate theories and find
their {\em propositional} models.

Let us observe that the algorithm described above can be implemented as a {\em
two-tier backtracking search}, with the on-line computation of supports of
guarded resolution proofs, 
and the usual backtracking scheme of DPLL. This second
backtracking can be implemented using any DPLL-based SAT-solver.
Proposition \ref{p.subeq} implies that the algorithm we outlined is both sound
and complete. Indeed,
if the SAT solver returns a model $M$ of theory $T$, then, by Proposition
\ref{p.subeq}(1), $M$ is a stable model for $P$. Otherwise we generate another 
candidate theory and loop through this process until one satisfying 
assignment is found.
Proposition \ref{p.subeq}(2) guarantees the completeness of our algorithm.

Our algorithm is not using loop formulas like the algorithm of Lin and Zhao
\cite{lz02} but systematically searches for supports of proof schemes, thus
providing supports for atoms in the putative model.  It also differs from the
modified loop formulas approach of Ferraris, Lee and Lifschitz \cite{fll06} in
that we do not consider loops of the call-graph of $P$ at all. Instead, we
compute systematically proof schemes and their supports for atoms.
While the time-complexity of our algorithm is significant because 
there may be exponentially many supports for any given atom 
$p$, the space complexity
is $|P|$. This is the effect of not looking at loop formulas at all
(\cite{lr06}).

\section{Conclusions and Further Work}\label{sec-concls}

We have shown that guarded unit resolution, 
a proof system that is a nonmonotonic
version of unit resolution, 
is adequate for description of the Gelfond-Lifschitz
operator $\gl_P$ and its fixpoints. That is, we can characterize stable models
of logic programs in terms of guarded resolution. 

There are several natural questions concerning extensions of guarded 
resolution in the context of  Answer Set Programming.  
For example:\\
\ \\
(1) {\em Is there an analogous proof
system for the disjunctive version of logic programming}? \\
or\\
(2) {\em Are there analogous
proof systems for logic programming with constraints?} \\
\ \\
We believe that
availability of such proof systems could help with finding a variety of results
on the complexity of reasoning in nonmonotonic logics. An interesting case is
that of propositional Default Logic. We will show in a subsequent paper that 
that we can develop a similar guarded resolution proof scheme 
for propositional Default Logic. The main difference is that we must 
allow proof trees where the leaves can be tautologies rather 
than just guarded atoms or guarded clauses that are derived from 
the given program $P$ as in this paper. 

\bibliographystyle{acmtrans}


\bibliography{guardRes0111}

\end{document}